\pdfoutput=1

\documentclass[11pt]{article}

\usepackage[final]{acl}

\usepackage{times}
\usepackage{latexsym}
\usepackage{amsmath}

\usepackage[T1]{fontenc}

\usepackage[utf8]{inputenc}

\usepackage{microtype}

\usepackage{inconsolata}

\usepackage{graphicx}

\usepackage{multirow}
\usepackage[inline]{enumitem}
\usepackage{scrextend}
\usepackage{array}
\usepackage{arydshln}

\usepackage{hhline}
\usepackage{subcaption}
\usepackage{hyperref}

\usepackage{booktabs, multirow} 
\usepackage{soul}
\usepackage{xcolor,colortbl} 
\usepackage{changepage,threeparttable} 

\newcommand{\squishlist}{
\begin{list}{$\bullet$}
{   \setlength{\itemsep}{0pt}
   \setlength{\parsep}{3pt}
   \setlength{\topsep}{3pt}
   \setlength{\partopsep}{0pt}
   \setlength{\leftmargin}{1.5em}
   \setlength{\labelwidth}{1em}
   \setlength{\labelsep}{0.5em} } }

\makeatletter
\newcommand*{\citelinktext}[2]{%
  \hyper@@link[cite]{}{cite.#1}{#2}}
\makeatother
   
\usepackage{amsmath}

\title{SemEval-2025 Task 1: AdMIRe - Advancing Multimodal Idiomaticity Representation}

\author{
Thomas Pickard\textsuperscript{1},
\textbf{Aline Villavicencio\textsuperscript{1,2}}
Maggie Mi\textsuperscript{1},
Wei He\textsuperscript{2},
Dylan Phelps\textsuperscript{1}
\and
\textbf{Marco Idiart\textsuperscript{3}} 
\\[0.3cm]
\textsuperscript{1} University of Sheffield, UK \\
\textsuperscript{2} University of Exeter, UK \\
\textsuperscript{3} Federal University of Rio Grande do Sul, Brazil \\ 
\texttt{\small \{tmrpickard1, zmi1, drsphelps1\}@sheffield.ac.uk} \\
\texttt{\small \{a.villavicencio, w.he\}@exeter.ac.uk, marco.idiart@gmail.com} 
}

\begin{document}
\maketitle
\begin{abstract}
Idiomatic expressions present a unique challenge in NLP, as their meanings are often not directly inferable from their constituent words. Despite recent advancements in Large Language Models (LLMs), idiomaticity remains a significant obstacle to robust semantic representation. 
We present datasets and tasks for SemEval-2025 Task 1: AdMIRe (Advancing Multimodal Idiomaticity Representation), which challenges the community to assess and improve models' ability to interpret idiomatic expressions in multimodal contexts and in multiple languages. Participants competed in two subtasks: ranking images based on their alignment with idiomatic or literal meanings, and predicting the next image in a sequence. The most effective methods achieved human-level performance by leveraging pretrained LLMs and vision-language models in mixture-of-experts settings, with multiple queries used to smooth over the weaknesses in these models' representations of idiomaticity.

\end{abstract}

\section{Introduction}

Idioms are a class of multi-word expression (MWE) which pose a challenge for current state-of-the-art models because their meanings are often not predictable from the individual words that compose them \cite{dankers-etal-2022-transformer,VILLAVICENCIO2005365}.  For instance, ``eager beaver" is unlikely to refer to a passionate muskrat; rather, it typically describes a person who is keen and enthusiastic. These expressions may also generate ambiguity between the literal, surface meaning arising from their component words and their idiomatic meaning \citep{he2024investigating}. These, among other characteristics, make them a valuable testing ground for examining how NLP models capture meaning.

Advances in language modeling of such phenomena have been made in recent years \citep{zeng-bhat-2022-getting, zeng-etal-2023-iekg, he-etal-2024-enhancing}, but large language models (LLMs) which perform well on general benchmarks (even for well-resourced languages such as English) fail to consistently exhibit good understanding of figurative language \citep{mi_rolling_2024, phelps-etal-2024-sign}.
This has an impact on their application in natural language processing activities such as sentiment analysis \citep{williams_role_2015,spasic_idiom-based_2020}, understanding and inference tasks and machine translation \citep{yazdani-etal-2015-learning,syahrir_analysis_2021}. For example, due to poor automatic translation of an idiom, the Israeli PM appeared to call the winner of Eurovision 2018 a `real cow' instead of a `real darling'!\footnote{\href{https://metro.co.uk/2018/05/13/israel-eurovision-winner-netta-called-a-real-cow-by-prime-minister-in-auto-translate-fail-7541925/}{metro.co.uk}}.

Several benchmark datasets exist for the processing of idiomatic expressions in text \citep[e.g.][]{FLUTE, haagsma_MAGPIE, ID10M, tayyar-madabushi-etal-2021-astitchinlanguagemodels-dataset, garcia-etal-2021-assessing, mi_rolling_2024} but concerns have been raised that these tasks do not necessarily require that language models possess good representations of idiom meaning \cite{boisson_constructionartifacts, he2024investigating}.
More recent datasets \cite{IRFL, VFLUTE_pp} introduce a visual modality to idiom processing tasks alongside text, and their findings indicate that this task is indeed more difficult for vision-language models (VLMs) to perform.

The task presented here builds on previous SemEval tasks exploring the evaluation of compositional models~\cite{marelli-etal-2014-semeval}, paraphrases and interpretation of noun compounds~\cite{hendrickx-etal-2013-semeval, butnariu-etal-2009-semeval} and idiomaticity detection~\cite{semeval-2022-task2}. We incorporate visual (\S\ref{sec:subtaskA_desc}) and visual-temporal (\S\ref{sec:subtaskB_desc}) modalities across two subtasks in an effort to promote the construction of higher-quality semantic representations of idioms. Our dataset incorporates items in both English (EN) and Brazilian Portuguese (PT-BR), and we focus on nominal compounds which have interpretations in literal and idiomatic senses which are both plausible and imageable.

The AdMIRe datasets are available from \href{http://doi.org/10.15131/shef.data.28436600.v1}{doi.org/10.15131/shef.data.28436600.v1} under a CC-BY-4.0 license.

\section{Dataset Construction}
\label{sec:dataset}

\subsection{Target compound selection} 

The potentially idiomatic noun compounds used in this study were sourced from existing datasets including NCTTI \cite{tayyar-madabushi-etal-2021-astitchinlanguagemodels-dataset}, FLUTE \cite{FLUTE} and MAGPIE \cite{haagsma_MAGPIE}, or identified by the researchers during the project. For the purposes of this task, we filtered out compositional expressions (e.g. \textit{olive oil}), focusing exclusively on those that exhibit duality — i.e., expressions that can be interpreted either literally or figuratively depending on the context (e.g. \textit{silver bullet}).
The candidate lists covered expressions in both English (EN) and Brazilian Portuguese (PT-BR).

The candidate expressions were then used to create two data subsets.

\subsection{Static Images (Subtask A)}
\label{sec:data_static}

Native speakers of the target language were asked to write a short sentence describing a visual scene depicting each target expression in the following contexts:

\begin{itemize}[nosep]
    \item strongly figurative
    \item mildly figurative
    \item mildly literal
    \item strongly literal
\end{itemize}
\vspace{0.5em}

In addition, a `distractor' prompt was also requested - this was something unrelated to either the figurative or literal meaning of the expression.
Where the annotators were unable to construct any of these scenes, the item was excluded as a candidate. For instance, it is difficult to capture the semantics of an idiomatic \textit{kangaroo court} in an image. Candidate items therefore needed to have plausible and imageable literal and idiomatic interpretations.

For each item selected, the annotators also provided two context sentences; one in which the expression is used literally, and one idiomatic. These sentences were obtained from existing corpora \cite{tayyar-madabushi-etal-2021-astitchinlanguagemodels-dataset,ententen-corpus} or were written specifically for AdMIRe.
For Portuguese, we observed that many adjective-noun compounds would normally be distinguished in writing between literal and idiomatic cases since they would be hyphenated in the idiomatic instance. For instance, \textit{ovelha-negra} means ``black sheep" (with the same idiomatic sense as English), but \textit{ovelha negra} would be a literal black sheep.  To avoid creating a shortcut for the models, we removed these hyphens from the idiomatic context sentences, and a native speaker reviewed them to confirm that they still appeared to be natural.

In total, we obtained 100 English and 55 Portuguese potentially idiomatic expressions, with literal and idiomatic context sentences and visual scene descriptions for each item.

\begin{table}[!htp]\centering
\caption{Data Sources for Static Images. }\label{tab: }
\scriptsize
\begin{tabular}{lrrr}\toprule
Source & EN & PT-BR \\\midrule
NCTTI & 54 & 54 \\
MAGPIE & 11 & - \\
Crowdsourced & 31 & 1 \\
FLUTE & 4 & - \\
\midrule
Total & 100 & 55 \\
\bottomrule
\end{tabular}
\end{table}

\begin{figure*}[ht!]

\begin{subfigure}[t]{0.19\textwidth}
  \centering
  \caption*{strongly figurative}
  \includegraphics[width=\textwidth]{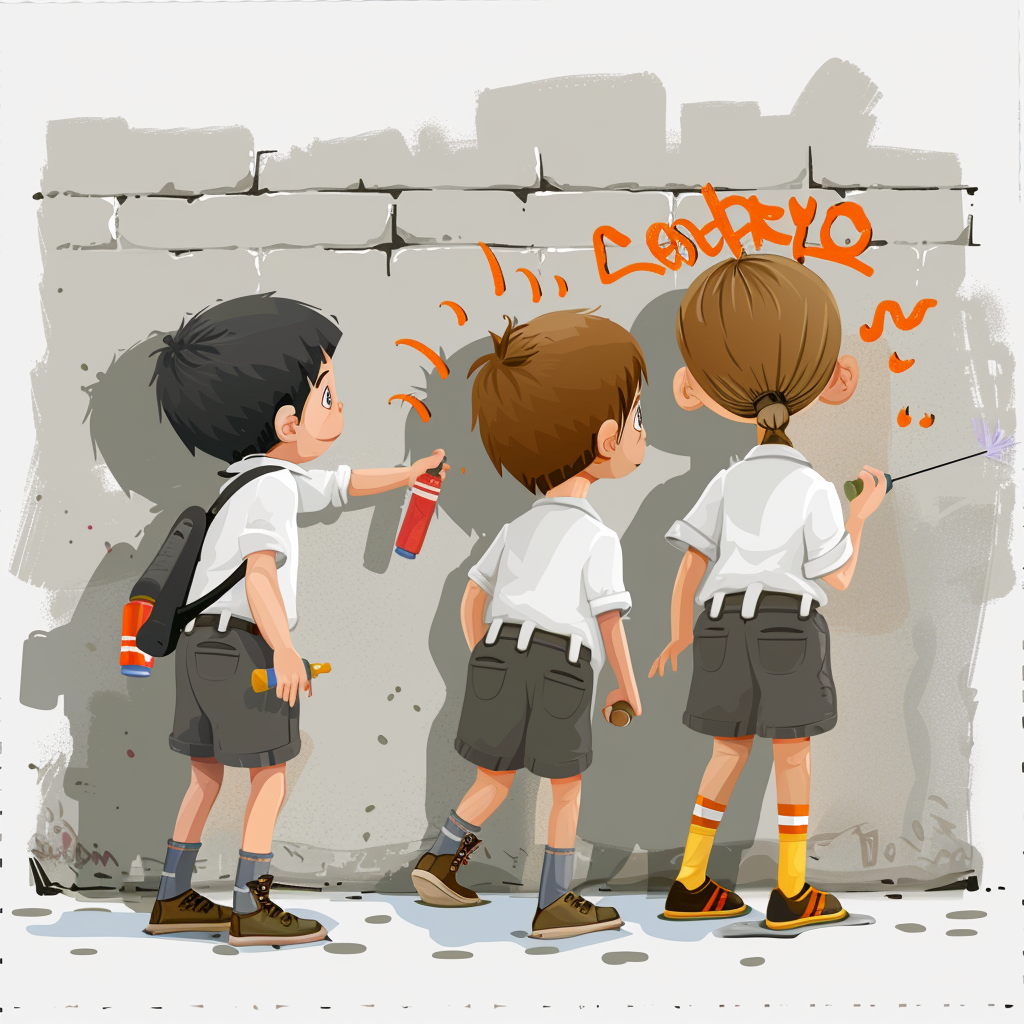}
  \caption{\centering The image depicts three children standing in front of a gray, textured wall... }
  \label{fig:ba1}
\end{subfigure}
 \hfill 
\begin{subfigure}[t]{0.19\textwidth}
  \centering
  \caption*{mildly figurative}
  \includegraphics[width=\textwidth]{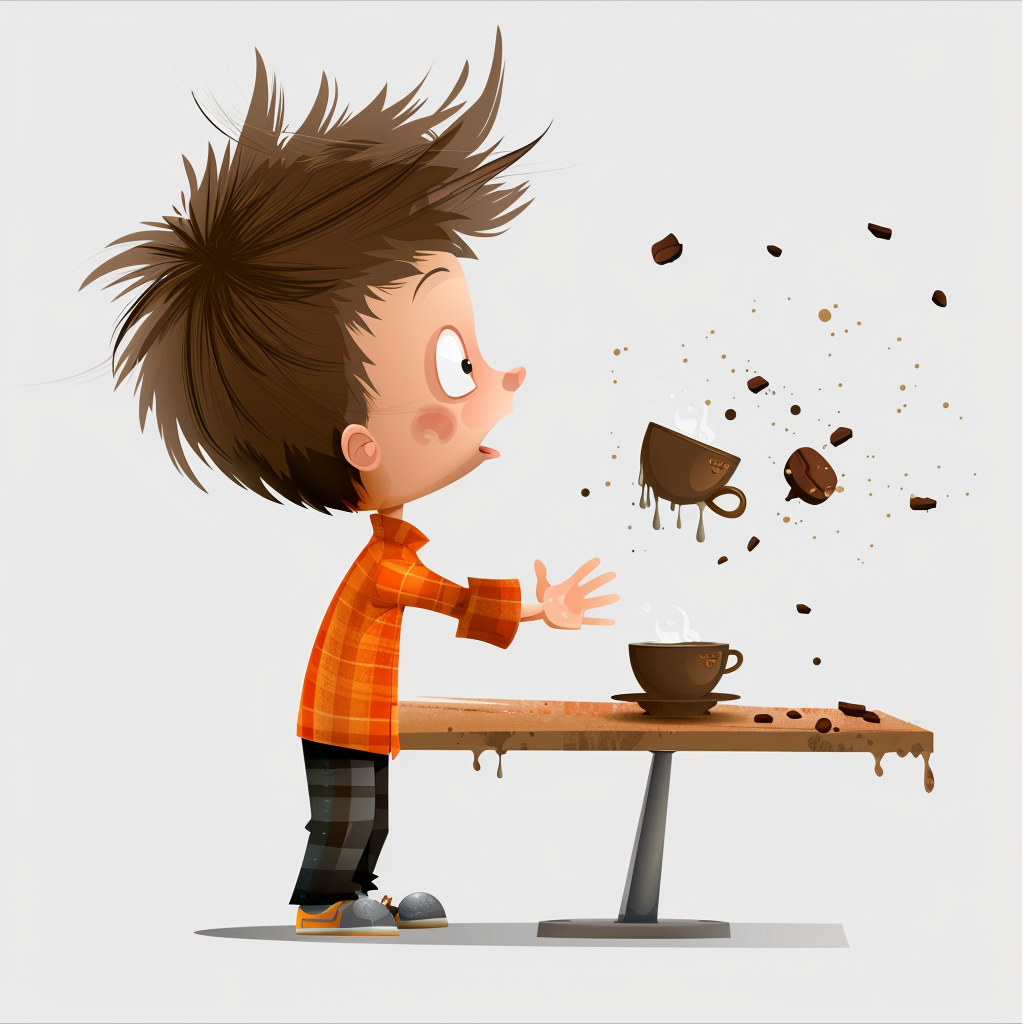}
  \caption{\centering The image depicts a cartoon-style illustration of a young boy standing at a table... }
  \label{fig:ba2}
\end{subfigure}
 \hfill 
\begin{subfigure}[t]{0.19\textwidth}
  \centering
  \caption*{distractor}
  \includegraphics[width=\textwidth]{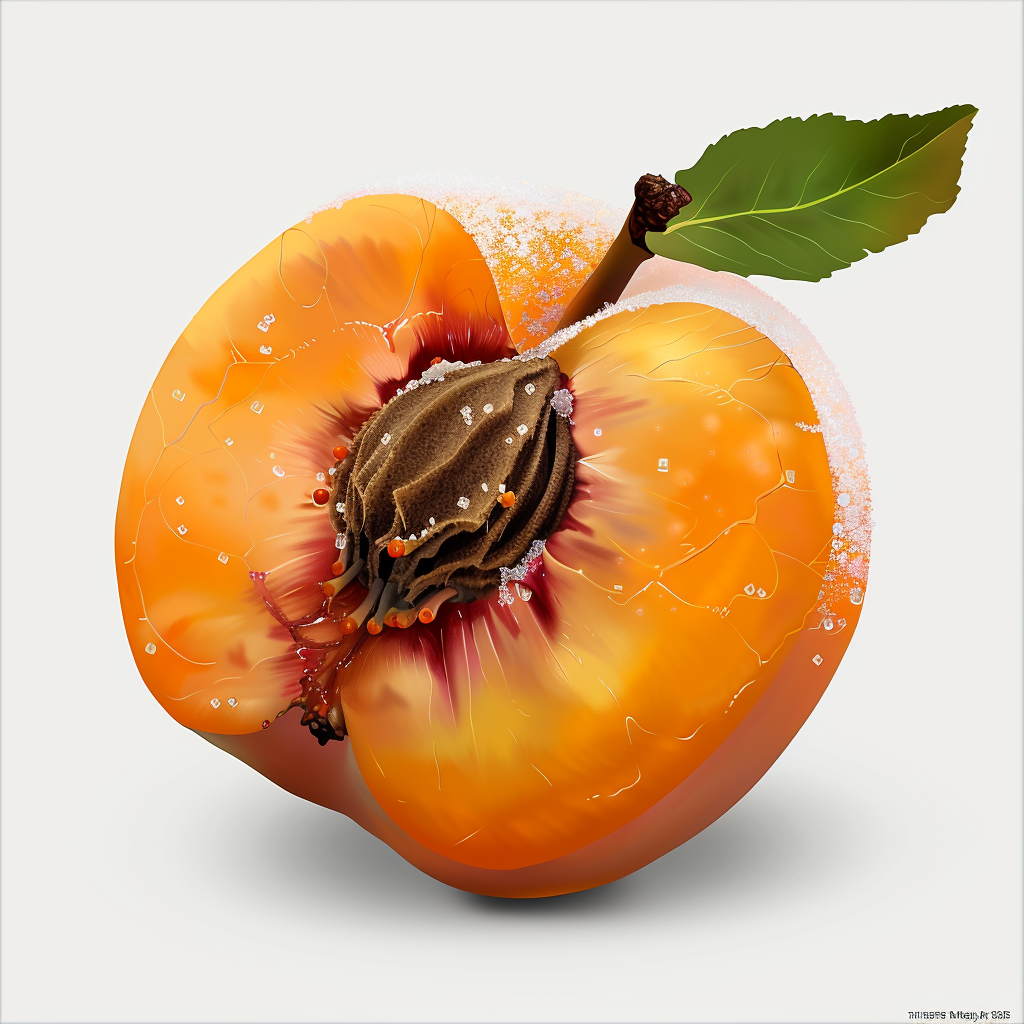}
  \caption{\centering The image depicts a halved peach with a detailed and realistic appearance...}
  \label{fig:ba3}
\end{subfigure}
\begin{subfigure}[t]{0.19\textwidth}
  \centering
  \caption*{mildly literal}
  \includegraphics[width=\textwidth]{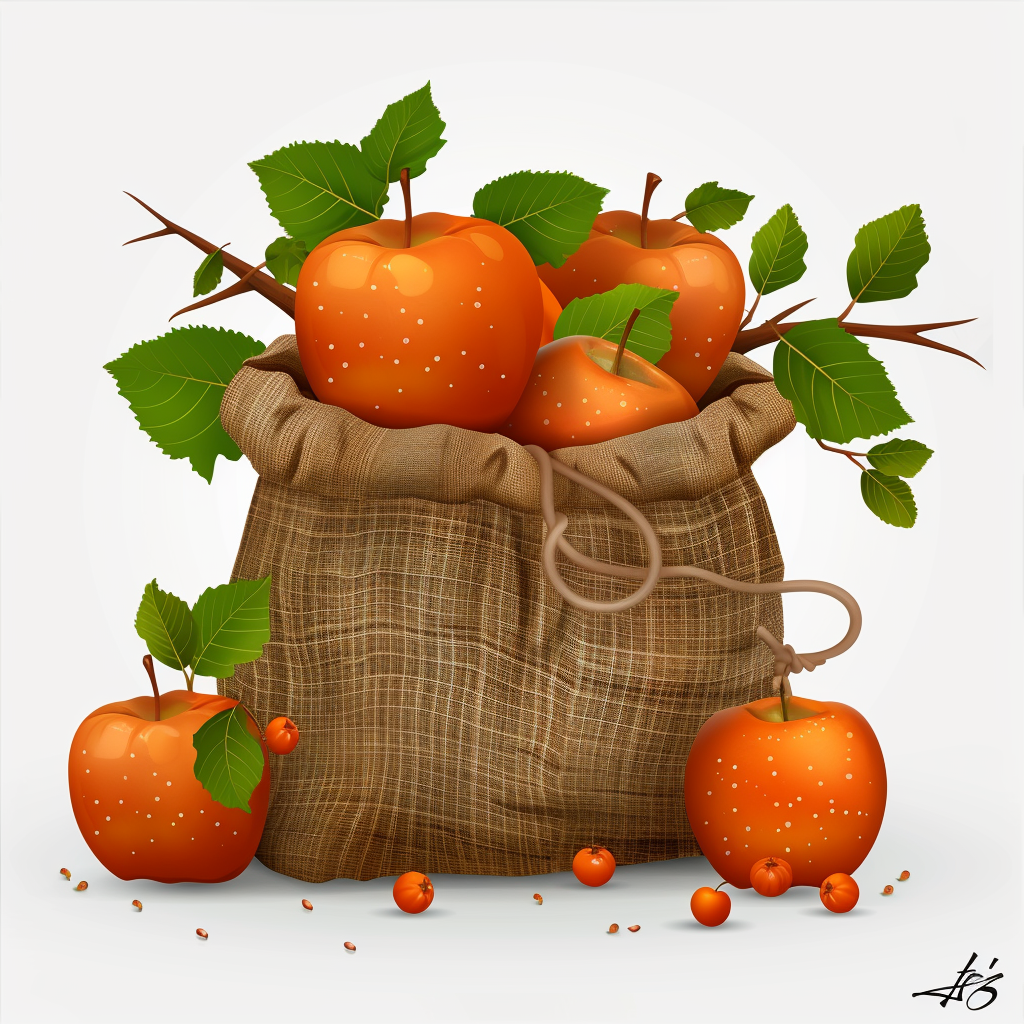}
  \caption{\centering The image depicts a rustic, burlap sack filled with several bright orange apples...}
  \label{fig:ba5}
\end{subfigure}
 \hfill 
\begin{subfigure}[t]{0.19\textwidth}
  \centering
  \caption*{strongly literal}
  \includegraphics[width=\textwidth]{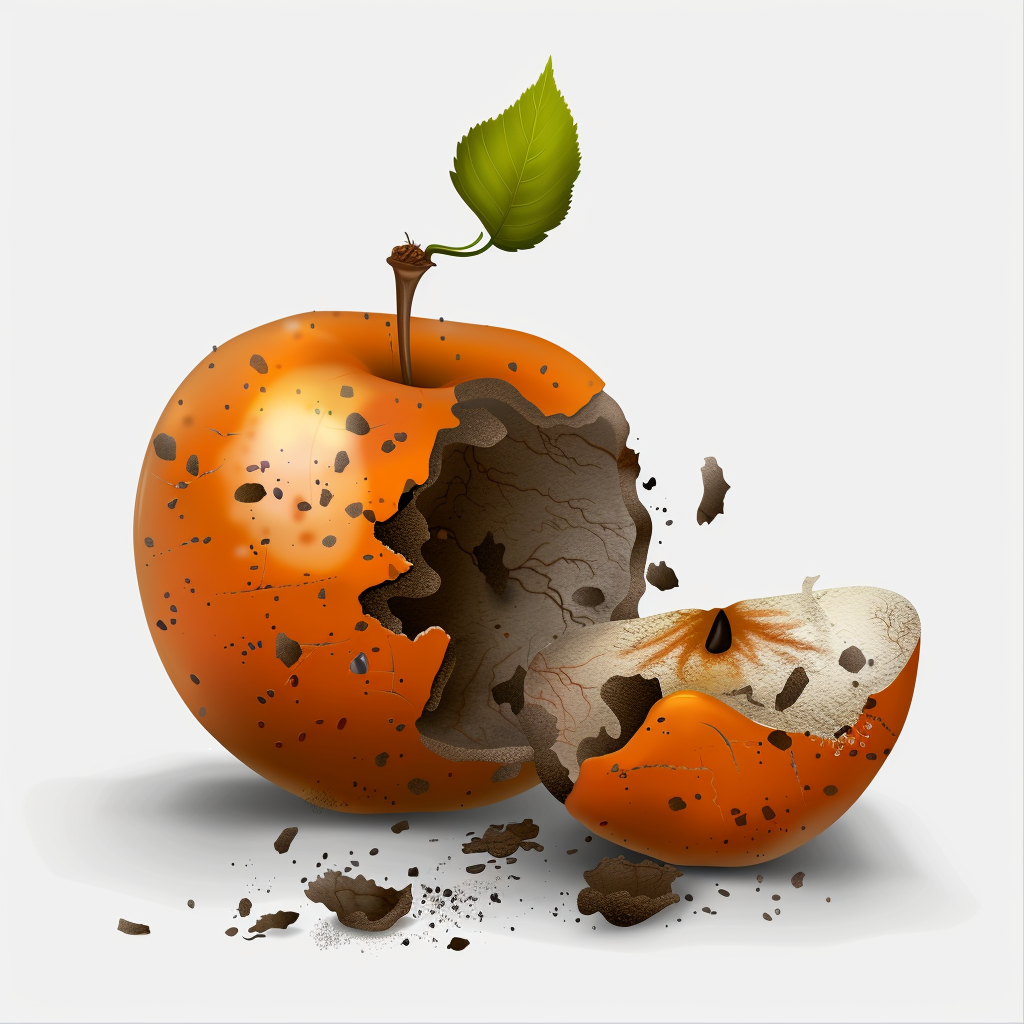}
  \caption{\centering The image depicts an orange-colored apple that appears to be decomposing or decaying...}
  \label{fig:ba4}
\end{subfigure}
 \hfill 


\captionsetup{skip=1.5em} 
\caption{\footnotesize \centering Subtask A data example for \textit{bad apple}. Images generated using Midjourney. Captions are displayed partially.}
    \label{fig:bad_apple_task_a}
\end{figure*}
\vspace{-0.5em}

\subsection{Image Sequences (Subtask B)}
\label{sec:data_sequence}

The semantics of many idiomatic expressions are difficult to capture in a single, static image, as they incorporate aspects of ongoing actions or changes over time. For instance, a \textit{brain drain} is not an instantaneous event but something which takes place over a period of time. 
In order to represent some of these items in the AdMIRe dataset, we also incorporated a visual-temporal modality.

For the image sequences dataset, native speakers of English were asked to write short descriptions of three visual scenes which form a sequence representing either the literal or idiomatic sense of a given expression, akin to a three-panel comic strip. For each of the sequences, they also wrote two alternatives to the final image in the sequence. These alternatives included elements relating to the previous panels but were incompatible with the target expression, and were intended to increase the difficulty of identifying the correct completion for the sequence based solely on image similarity without using the semantic information.

A total of 30 items were collected in English. Note that 10 items appear in both of the English data subsets.

\subsection{Image Generation}

For each visual scene created by the annotators, we used a commercial text-to-image diffusion model (Midjourney\footnote{\url{https://www.midjourney.com/}}) to generate a corresponding image. This choice of tool provided the fine-grained human control needed to produce high-quality, domain-specific images tailored to the task, guided by human expert supervision to ensure  alignment with literal and idiomatic meanings.
A consistent `style reference' image was used, guiding the model to produce images with a consistent, cartoon-like appearance. For image sequences where the same character(s) were needed, we also employed a set of character reference images in the same style.




\subsection{Caption Generation}
\label{sec:caption}

In order to support participation in the shared task by teams who wish to work only with text models (which lowers complexity and computational costs), we generated descriptive captions for each image. These were obtained by using the {\emph{LLaVA-HF/v1.6-mistral-7b-hf}}\footnote{https://huggingface.co/llava-hf/llava-v1.6-mistral-7b-hf} (LLaVA) model, a large vision-language model specifically designed for tasks requiring multimodal reasoning \citep{liu2024visual}. LLaVA integrates a vision encoder to extract semantic features from images and a large language model to process these features and generate text. By employing the prompt ``\emph{What is shown in this image?}'', the model generates captions that describe the content of the input images. The workflow ensures that the visual and textual components of the model work in harmony to produce accurate and contextually relevant descriptions. To ensure the quality of the generated captions, all outputs were reviewed and verified by human evaluators. 
For items in Portuguese, we provide captions in both English and Portuguese.

\section{Task Description}
\label{sec:task}

\begin{figure*}[ht!]
\hspace{1.2cm}\small{First two of the sequence:} 
\hspace{5.4cm}\small{Candidates:}
\vspace{0.1em}

\begin{subfigure}[t]{0.16\textwidth}
  \centering
  \includegraphics[width=\textwidth]{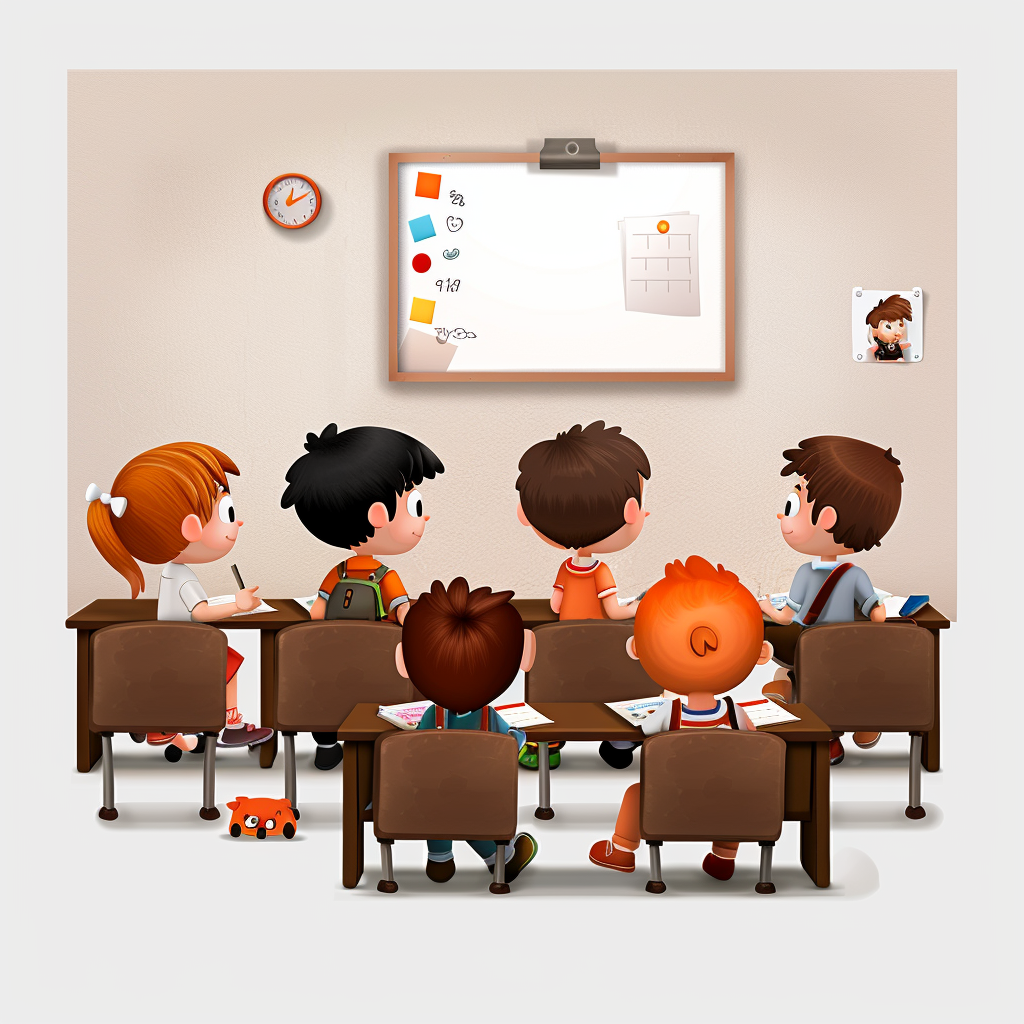}
  \caption{\centering \scriptsize The image shows a classroom scene with five animated characters, likely children, sitting at desks...}
  \label{fig:ba_strip1}
\end{subfigure}
\begin{subfigure}[t]{0.16\textwidth}
  \centering
  \includegraphics[width=\textwidth]{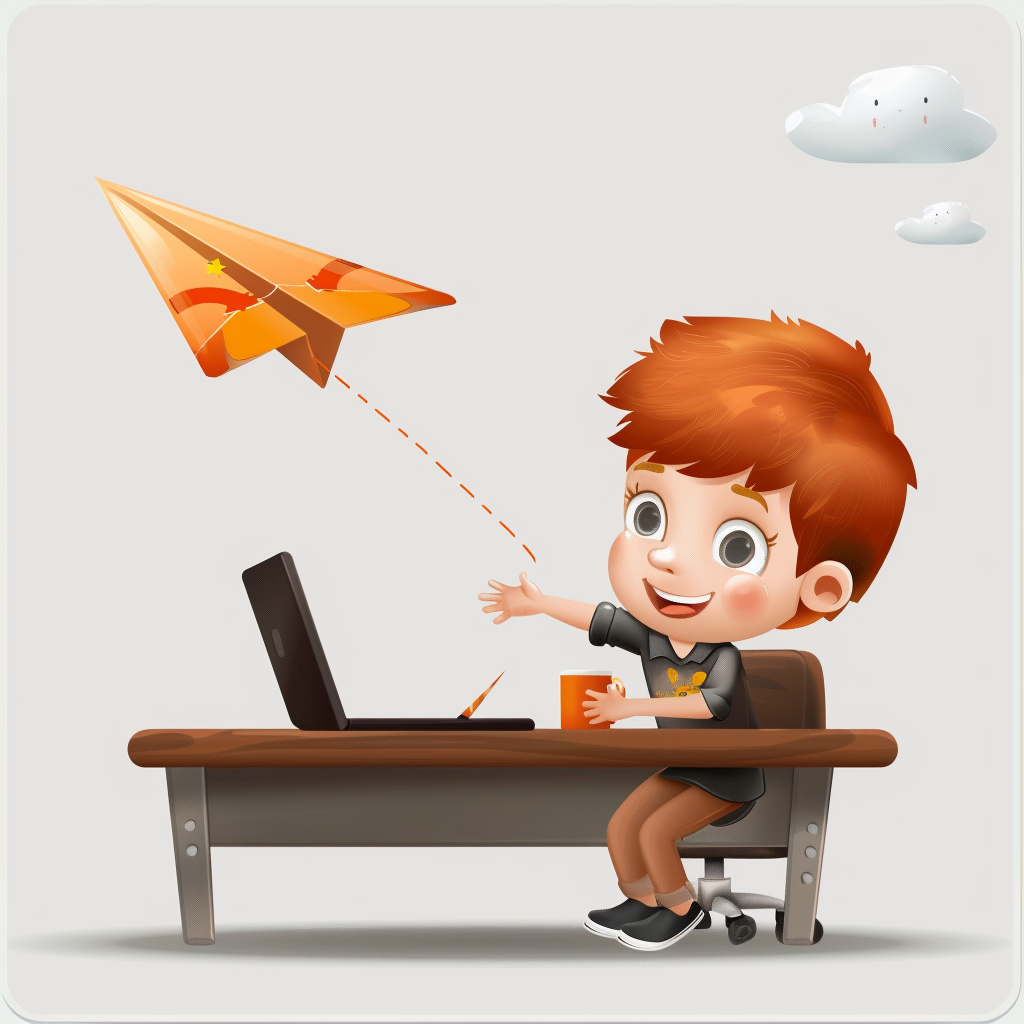}
  \caption{\centering \scriptsize The image shows an animated character, a young boy with red hair, sitting at a desk with a laptop...}
  \label{fig:ba_strip2}
\end{subfigure}
 \hspace{0.015\textwidth}
 \begin{subfigure}[t]{0.16\textwidth}
  \centering
  \includegraphics[width=\textwidth]{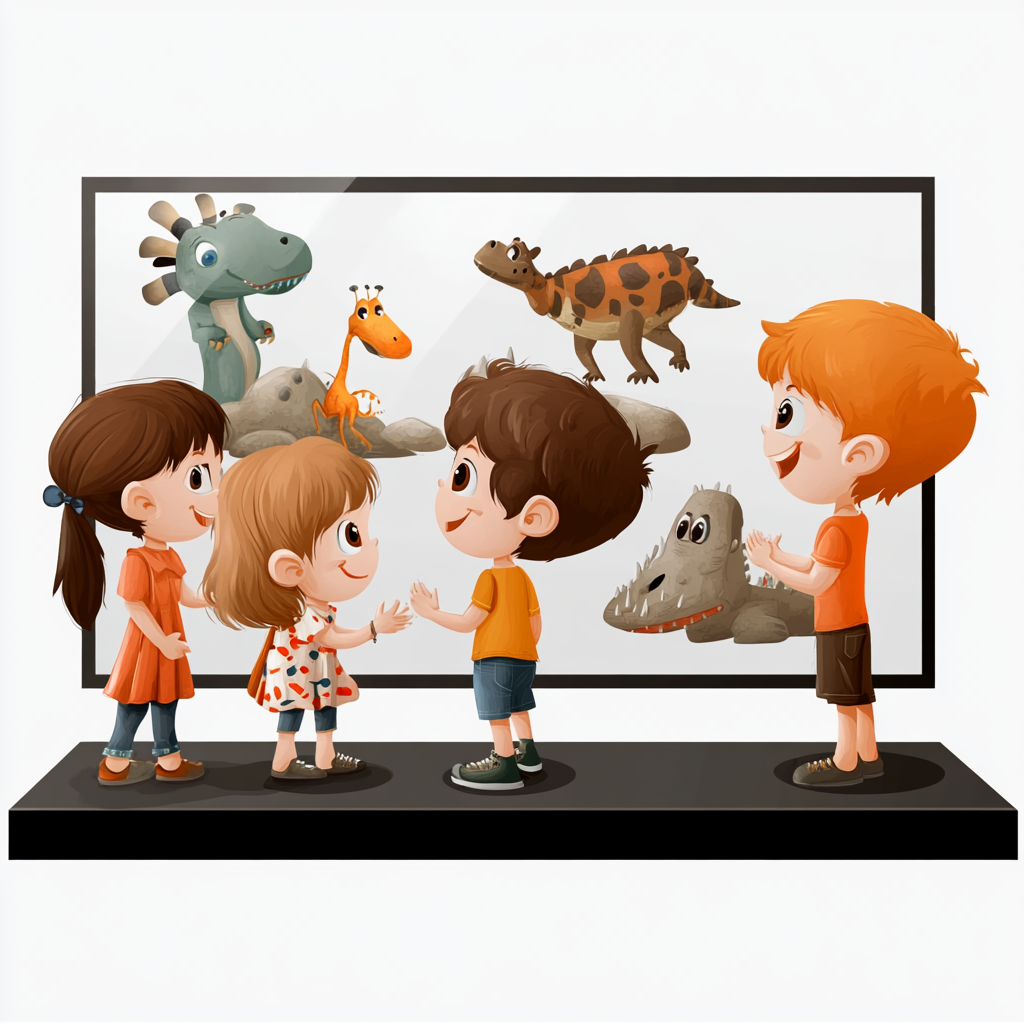}
  \caption{\centering \scriptsize The image shows a group of animated characters, likely children, standing in front of a television screen that displays various cartoon animals...}
  \label{fig:ba_strip3}
\end{subfigure}
\begin{subfigure}[t]{0.16\textwidth}
  \centering
  \includegraphics[width=\textwidth]{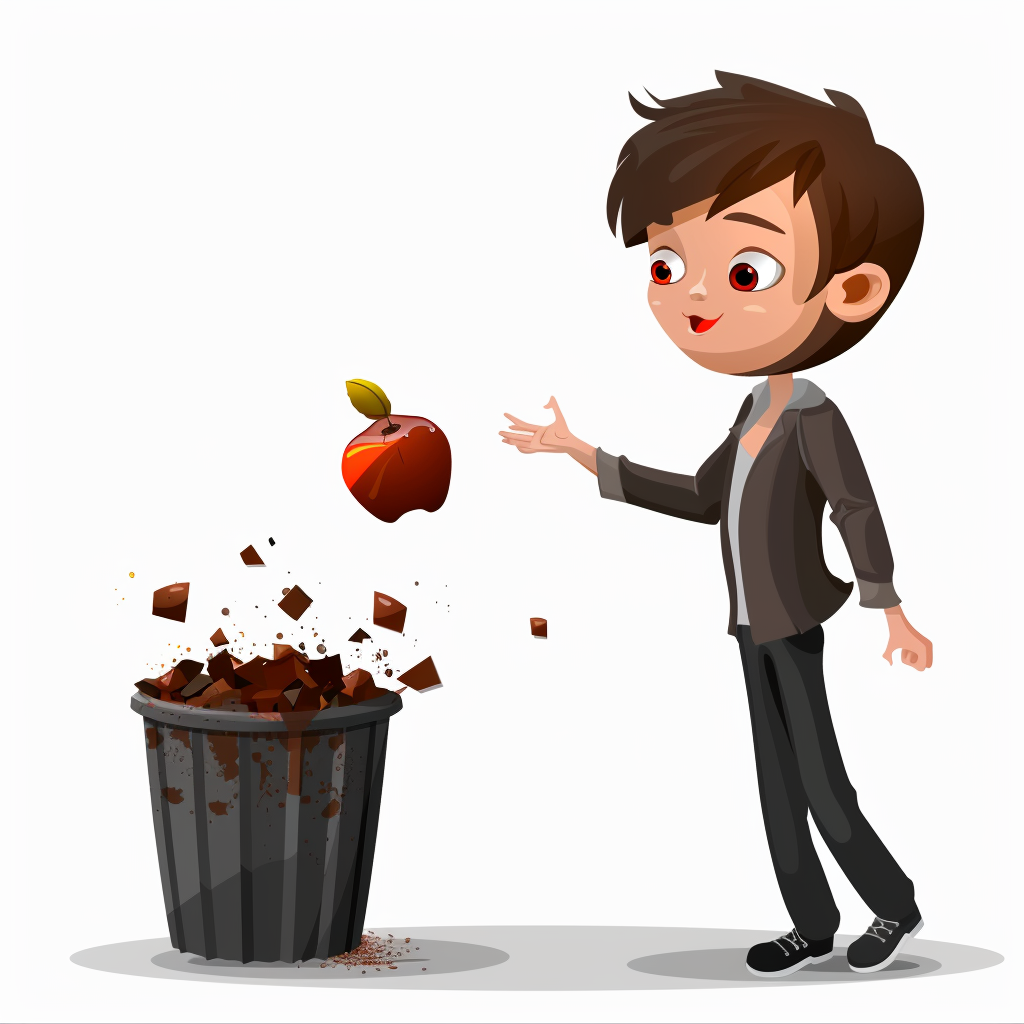}
  \caption{\centering \scriptsize The image shows an animated character, a young man with brown hair, standing next to a trash can...}
  \label{fig:ba_strip4}
\end{subfigure}
\begin{subfigure}[t]{0.16\textwidth}
  \centering
  \includegraphics[width=\textwidth]{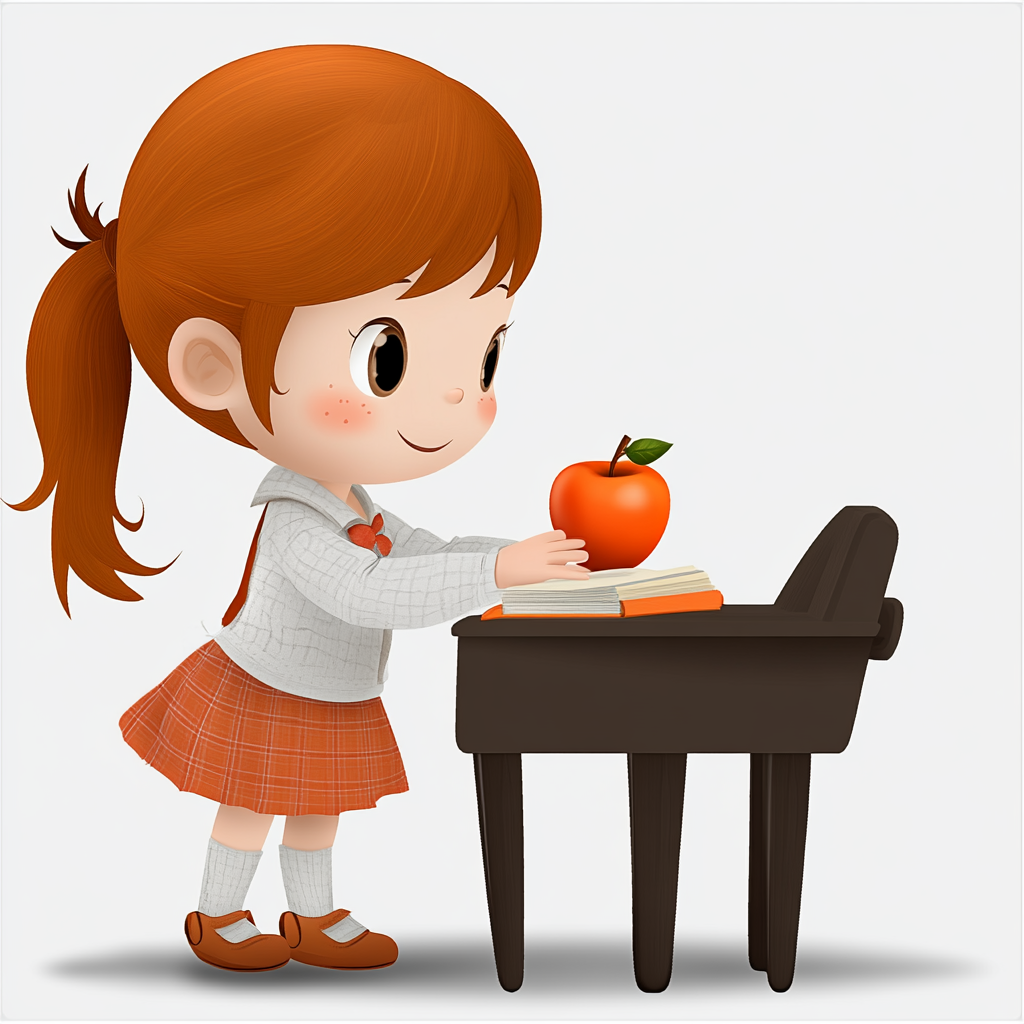}
  \caption{\centering \scriptsize The image shows a cartoon illustration of a young girl with red hair, wearing a white sweater with a red bow and a plaid skirt...}
  \label{fig:ba_strip5}
\end{subfigure}
 \begin{subfigure}[t]{0.16\textwidth}
  \centering
  \includegraphics[width=\textwidth]{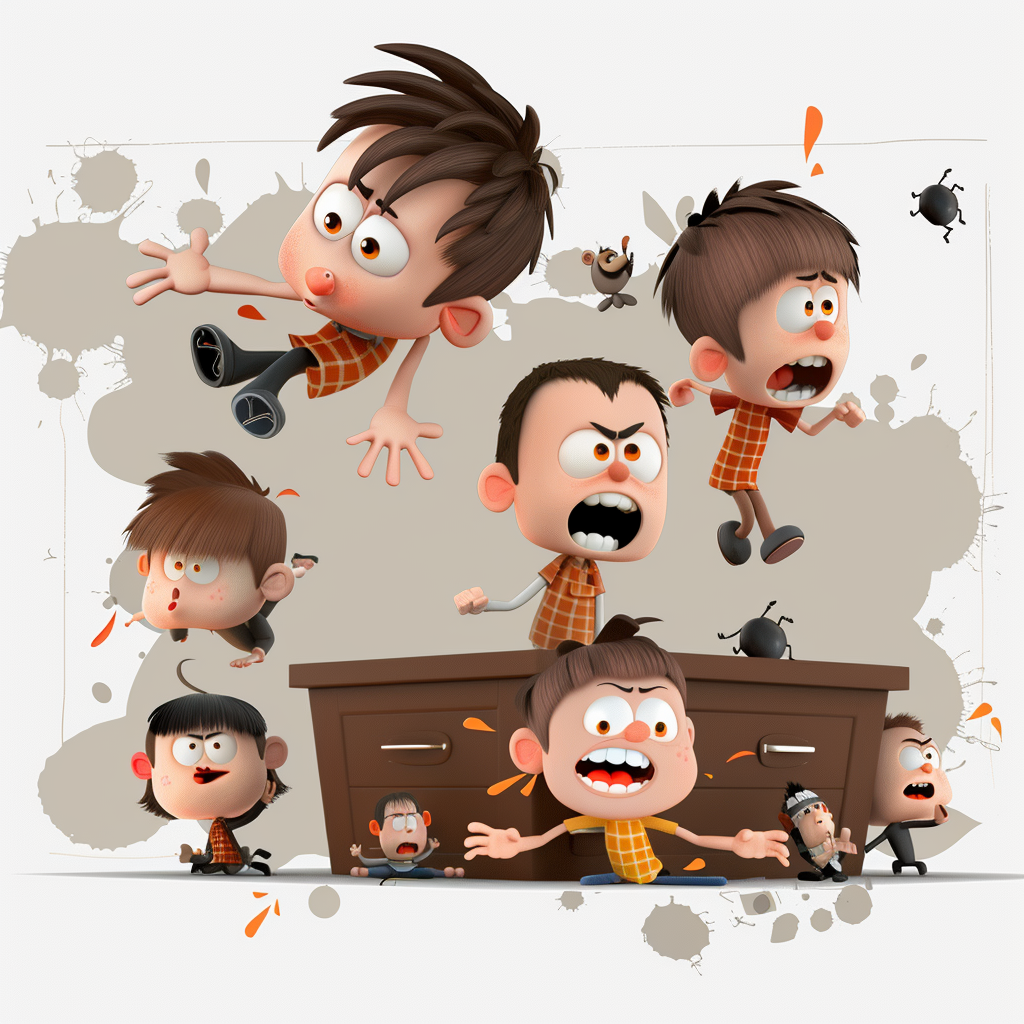}
  \caption{\centering \scriptsize The image shows a group of animated characters that appear to be in a state of distress or chaos...}
  \label{fig:ba_strip6}
\end{subfigure}

\captionsetup{skip=1.5em} 
\caption{\footnotesize  \centering Subtask B data example for \textit{bad apple}.  
Images (a) and (b) form the initial part of the sequence, while images (c) through (f) serve as completion candidates. In this instance, the intended sense of \textit{bad apple} is idiomatic.
}    
    \label{fig:bad_apple_task_b}
\end{figure*}

\subsection{Task A: Multiple Image Choice}
\label{sec:subtaskA_desc}

Subtask A uses the static image portion of the AdMIRe dataset (\S\ref{sec:data_static}). Given a context sentence containing a potentially idiomatic nominal compound (NC) and a set of five images, the task is to \textbf{rank} the images based on how accurately they depict the meaning of the NC used in that sentence. 

A variation of task also allows for monomodal settings, where given a sentence and five text captions (each describing the content of one of the images, as described in Section~\ref{sec:caption}) the goal is to rank the image captions on how they capture the meaning of the NC.

Figure \ref{fig:bad_apple_task_a} provides an example of the Subtask A data for the expression \textit{bad apple}.

The dataset is split into training, development and test sets, with each compound present in only one subset and one, randomly-selected, sense (literal or idiomatic).
In addition, we provide an extended evaluation set, which uses all 100 compounds (and therefore overlaps with the other subsets), but selects the other sense of the NC. For example, \textit{elbow grease} appears with its idiomatic sense in the training dataset, and literally in the extended evaluation set.

The English dataset for Subtask A includes 70 training items (350 image-caption pairs), 15 development items, and 15 test items, while the Portuguese dataset comprises 32 training items (160 image-caption pairs), 10 development items, and 13 test items.

\subsection{Subtask B: Image Sequences (Next Image Prediction)}
\label{sec:subtaskB_desc}


Subtask B uses the image sequence portion of the AdMIRe dataset (\S\ref{sec:data_sequence}).

For each target expression, the first two images in either the literal or idiomatic sequence are provided, along with a set of four candidate images for completion of the sequence. The candidates consist of the intended completion, the two associated alternatives and the completion for the other sense (literal/idiomatic) of the NC.
The task is to correctly identify the intended completion image, while also determining whether the depicted sense of the nominal compound (NC) is idiomatic or literal. Examples are shown in Figure \ref{fig:bad_apple_task_b}.

As with Subtask A, we also offer two settings for Subtask B, with descriptive text replacing the images in the `caption' setting. In the Subtask B dataset, the English set includes 20 examples for training, 5 for development, and 5 for testing. All 30 items are also included in an extended evaluation set, with initial sequences and completion candidates appropriate to the NC sense not used in the primary data.

\subsection{Evaluation}

\subsubsection{Subtask A}

For subtask A, we set an expected rank ordering of the 5 images which depends on the sense in which the expression is used in the context sentence. The image strongly associated with the target sense should be ranked first, followed by the mildly associated one. The images for the other sense follow, while the `distractor' image is always expected to be least relevant.
For instance, if \textit{bad apple} is used idiomatically in the context sentence then the expected ranking for the images would be: strongly figurative, miildly figurative, mildly literal, strongly literal, distractor. For the images in Figure \ref{fig:bad_apple_task_a}, this would produce $[a,b,d,e,c]$.

Performance for Subtask A is assessed with two key metrics:
a) Top Image Accuracy, which measures only the correct identification of the \textbf{most} representative image and b) Discounted Cumulative Gain (DCG) \citep{jarvelin2002cumulated}, an established information retrieval metric that not only captures the fraction of retrieved relevant information but also takes into account their correct ordering.

\noindent
\textit{Discounted Cumulative Gain (DCG)} is defined as
\begin{equation*}
\label{eq:dcg}
\text{DCG}_n = \sum_{i=1}^{n} \frac{rel_i}{\log_2(i+1)},
\end{equation*}
where $rel_i$ is the relevance score of the $i$-th item, and $n$ is the number of items considered. 

Because our expected order of images is somewhat arbitrary (for a literal instance of a given expression, the idiomatic depictions are essentially no more relevant than the distractor), after experimentation we adopt a weighting of $[3,1,0,0,0]$ for the five image positions; this allows the metric to capture some of the relevant semantics beyond the top image accuracy without penalising systems which permute the order of the low-relevance images.
The maximum DCG score obtainable is therefore 3.631.

Competition rankings for Subtask A are based on top image accuracy, with DCG breaking ties.

\subsubsection{Subtask B}

This subtask assesses the model's ability to complete a sequence of images that narratively represent an idiomatic expression, along with distinguishing between idiomatic and literal meanings. Evaluation metrics for subtask B are a) Image completion accuracy, which measures the correctness of the selected image to complete the narrative and b) Sentence type accuracy, measuring the effectiveness in identifying idiomatic versus literal expressions.


\begin{table}[h]
    \centering
    \small
    \begin{tabular}{c|c|c|c}
         \textbf{Subtask} & \textbf{Language} & \multicolumn{2}{|c}{\textbf{Track}} \\
         & & Text \& Images &  Text-Only \\
         \midrule
         \multirow{2}{*}{A} & English & 16 & 5 \\
         & Portuguese & 11 & 1 \\
         \midrule
         B & English & 3 & 1 \\
         \hline
    \end{tabular}
    \caption{\centering Number of submissions made to each subtask}
    \label{tab:participants}
    \vspace{-1em}
\end{table}

\begin{table*}[ht]
    \centering
    \small
    \begin{tabular}{c|c|c|c|c|c|c}
        \toprule
        \multirow{2}{*}{Team} &  \multicolumn{3}{c}{Test Set Metrics} & \multicolumn{3}{c}{Extended Evaluation Metrics} \\
        & Rank & Top 1 Acc & DCG Score & Rank & Top 1 Acc & DCG Score \\
        \midrule
        \citelinktext{PALI-NLP}{PALI-NLP} & 1 & 0.93 & 3.52 & 1 & 0.83 & 3.43 \\
        \citelinktext{dutir914}{dutir914} & 2 & 0.93 & 3.46 & 3 & 0.79 & 3.28 \\
        \citelinktext{AlexUNLP-NB}{AlexUNLP-NB} & 3 & 0.93 & 3.45 & 5 & 0.72 & 3.22 \\
        \citelinktext{AIMA}{AIMA}     & 4 & 0.87 & 3.44 & 10 & 0.48 & 2.90 \\
        \citelinktext{daalft}{daalft}   & 5 & 0.87 & 3.43 & 2 & 0.81 & 3.35 \\
        \citelinktext{PoliTo}{PoliTo}	 & 6 &	0.87 & 3.381 & 4 & 0.75 & 3.20 \\
        \citelinktext{UCSC_NLP_T6}{UCSC NLP T6}	& 7 & 0.87 & 3.36 & - & - & - \\		
        \citelinktext{Zhoumou}{Zhoumou}	 & 8 & 0.73 & 3.20 & 6 & 0.69 & 3.21 \\
        \citelinktext{HiTZ-Ixa}{HiTZ-Ixa} & 9 & 0.73 & 3.13 & 7 & 0.58 & 3.00 \\
        \citelinktext{TueCL}{TueCL} & 10 & 0.67 &	3.16 & - & - & - \\
        \citelinktext{Howard_University-AI4PC}{Howard University-AI4PC} & 11 & 0.67 & 3.13 & - & - & - \\
        \citelinktext{UoR-NCL}{UoR-NCL}	 & 12 &	0.67 & 3.10 & 8 & 0.57 & 2.96 \\
        \citelinktext{FJWU_Squad}{FJWU\_Squad} & 13 & 0.60 & 2.90 & 11 & 0.47 & 2.85 \\
        \citelinktext{JNLP}{JNLP}     & 14 &	0.53 & 3.14 & 9 & 0.55 & 3.13 \\
        \citelinktext{Modgenix}{Modgenix} & 15 & 0.53 & 2.82 & - & - & - \\
        \citelinktext{YNU-HPCC}{YNU-HPCC} & 16 & 0.47 & 2.85 & - & - & - \\
        \citelinktext{UMUTeam}{UMUTeam}  & 17 & 0.40 & 2.68 & 12 & 0.24 & 2.52 \\
    \bottomrule

    \end{tabular}
    \caption{\centering Leaderboard results - \textbf{Subtask A, English}, Text \& Images}
    \label{tab:res_a_en_ti}
\end{table*}

\begin{table*}[]
    \centering
    \small
    \begin{tabular}{c|c|c|c|c|c|c}
        \toprule
        \multirow{2}{*}{Team} &  \multicolumn{3}{c}{Test Set Metrics} & \multicolumn{3}{c}{Extended Evaluation Metrics} \\
        & Rank & Top 1 Acc & DCG Score & Rank & Top 1 Acc & DCG Score \\
        \midrule
        \citelinktext{CTYUN-AI}{CTYUN-AI} & 1 & 0.87 & 3.51 & 1 & 0.64 & 3.10 \\
        \citelinktext{daalft}{daalft}   & 2 & 0.67 & 3.07 & 4 & 0.33 & 2.61 \\
        \citelinktext{JNLP}{JNLP}     & 3 & 0.67 & 3.04 & 3 & 0.51 & 2.86 \\
        \citelinktext{Transformer25}{Transformer25} & 4 & 0.47 & 2.82 & 2 & 0.54 & 3.04 \\
        \citelinktext{ChuenSumi}{ChuenSumi}   & 5 & 0.40 & 2.89 & 5 & 0.29 & 2.68 \\
        \bottomrule
    \end{tabular}
    \caption{\centering Leaderboard results - \textbf{Subtask A, English}, Text Only}
    \label{tab:res_a_en_to}
\end{table*}

\section{Participating Systems and Results}

\nocite{CTYUN-AI}
\nocite{ChuenSumi}
\nocite{daalft}
\nocite{UoR-NCL}
\nocite{AlexUNLP-NB}
\nocite{UMUTeam}
\nocite{Modgenix}
\nocite{Zhoumou}
\nocite{dutir914}
\nocite{PALI-NLP}

\nocite{JNLP}
\nocite{FJWU_Squad}
\nocite{Howard_University-AI4PC}
\nocite{PoliTo}
\nocite{YNU-HPCC}
\nocite{UCSC_NLP_T6}
\nocite{HiTZ-Ixa}
\nocite{AIMA}
\nocite{Transformer25}
\nocite{TueCL}


The AdMIRe shared task competitions were configured using the Codabench platform \cite{codabench}, with the main benchmark (Subtask A, images \& text) attracting 198 registered participants\footnote{We encountered a number of automated registrations which may have inflated this number somewhat. It is unclear what anyone hoped to achieve by generating spam registrations for the benchmark.}. 
Users were allowed to make multiple submissions during the competition, and were able to select their best result for evaluation. Submissions during the test phase (which determined the final leaderboard position) were limited to 5 in order to discourage 'gaming' the system while allowing participants to evaluate more than one approach if desired.

Once the competition ended, teams were asked to complete a brief questionnaire outlinging their approach and enabling us to link CodaBench usernames with team names in their system description papers.
Only teams who submitted a system description paper are included in the official task leaderboards. A total of 20 official team submissions were received. The number of submitted entries for each combination of subtask, track and language is summarised in Table \ref{tab:participants}.


\subsection{Results}

Results for each combination of subtask, language and track are presented in tables \ref{tab:res_a_en_ti} - \ref{tab:res_b_en_to}.


\begin{table*}[ht!]
    \centering
    \small
    \begin{tabular}{c|c|c|c|c|c|c}
        \toprule
        \multirow{2}{*}{Team} &  \multicolumn{3}{c}{Test Set Metrics} & \multicolumn{3}{c}{Extended Evaluation Metrics} \\
        & Rank & Top 1 Acc & DCG Score & Rank & Top 1 Acc & DCG Score \\
        \midrule
        \citelinktext{HiTZ-Ixa}{HiTZ-Ixa} & 1 & 1.00 & 3.51 & 7 & 0.45 & 2.82 \\
        \citelinktext{dutir914}{dutir914} & 2 & 0.92 & 3.43 & 2 & 0.69 & 3.06 \\
        \citelinktext{Zhoumou}{Zhoumou}  & 3 & 0.85 & 3.33 & 3 & 0.67 & 3.10 \\
        \citelinktext{daalft}{daalft}   & 4 & 0.77 & 3.31 & 4 & 0.56 & 2.95 \\
        \citelinktext{PALI-NLP}{PALI-NLP} & 5 & 0.69 & 3.21 & 1 & 0.76 & 3.23 \\
        \citelinktext{AlexUNLP-NB}{AlexUNLP-NB} & 6 & 0.62 & 3.09 & 6 & 0.51 &	2.91 \\
        \citelinktext{UoR-NCL}{UoR-NCL}	 & 7 & 0.54 & 3.05 & 5 & 0.56 &	2.90 \\
        \citelinktext{YNU-HPCC}{YNU-HPCC} & 8 & 0.38 & 2.92 & - & - & - \\
        \citelinktext{UMUTeam}{UMUTeam}	 & 9 & 0.38 & 2.57 & 8 & 0.18 &	2.33 \\
        \citelinktext{Howard_University-AI4PC}{Howard University-AI4PC} & 10 & 0.23 & 2.64 & - & - & - \\
        \citelinktext{Modgenix}{Modgenix} & 11 & 0.23 & 2.57 & - & - & - \\
        \bottomrule
    \end{tabular}
    \caption{\centering Leaderboard results - \textbf{Subtask A, Portuguese}, Text \& Images}
    \label{tab:res_a_pt_ti}
\end{table*}

\begin{table*}[]
    \centering
    \small
    \begin{tabular}{c|c|c|c|c|c|c}
        \toprule
        \multirow{2}{*}{Team} &  \multicolumn{3}{c}{Test Set Metrics} & \multicolumn{3}{c}{Extended Evaluation Metrics} \\
        & Rank & Top 1 Acc & DCG Score & Rank & Top 1 Acc & DCG Score \\
        \midrule
        \citelinktext{CTYUN-AI}{CTYUN-AI} & 1 & 0.92 & 3.43 & 1 & 0.56 & 2.97 \\
        \bottomrule
    \end{tabular}
    \caption{\centering Leaderboard results - \textbf{Subtask A, Portuguese}, Text Only}
    \label{tab:res_a_pt_to}
\end{table*}

\begin{table*}[]
    \centering
    \small
    \begin{tabular}{c|c|c|c|c|c|c}
        \toprule
        \multirow{2}{*}{Team} &  \multicolumn{3}{c}{Test Set Metrics} & \multicolumn{3}{c}{Extended Evaluation Metrics} \\
        & Rank & Image Accuracy & Sentence Type & Rank & Image Accuracy & Sentence Type \\
        \midrule
        \citelinktext{daalft}{daalft}   & 1 & 0.60 & 1.00 & 2 & 0.23 & 0.77 \\
        \citelinktext{PALI-NLP}{PALI-NLP} & 2 & 0.60 & 0.80 & 1 & 0.93 & 1.00 \\
        \citelinktext{Modgenix}{Modgenix} & 3 & 0.60 & 0.60 & - & - & - \\
    \bottomrule
    \end{tabular}
    \caption{\centering Leaderboard results - \textbf{Subtask B, English}, Text \& Images}
    \label{tab:res_b_en_ti}
\end{table*}

\begin{table*}[]
    \centering
    \small
    \begin{tabular}{c|c|c|c|c|c|c}
        \toprule
        \multirow{2}{*}{Team} &  \multicolumn{3}{c}{Test Set Metrics} & \multicolumn{3}{c}{Extended Evaluation Metrics} \\
        & Rank & Image Accuracy & Sentence Type & Rank & Image Accuracy & Sentence Type \\
        \midrule
        \citelinktext{daalft}{daalft}   & 1 & 1.00 & 1.00 & 2 & 0.60 & 0.77 \\
        \citelinktext{PALI-NLP}{PALI-NLP} & 2 & 0.80 & 0.60 & 1 & 0.60 & 0.90 \\
        \bottomrule
    \end{tabular}
    \caption{\centering Leaderboard results - \textbf{Subtask B, English}, Text Only}
    \label{tab:res_b_en_to}
\end{table*}

\subsection{Popular Approaches}

\paragraph{Model Types} Participating teams employed a variety of approaches to the AdMIRe subtasks. All teams opted to use prompting methods with large, generative language and vision-language models and/or to fine-tune smaller pretrained models on the task. Popular model families included GPT-4 \cite{GPT4card}; QWEN \cite{QWEN}; SBERT \cite{SBERT}; RoBERTa \cite{roberta}; CLIP \cite{clip} and its variants such as CLIP-ViLT \cite{ViLT-CLIP}, AltCLIP \cite{altclip} and ALIGN \cite{align}, and BLIP \cite{blip}. There was also some exploration of the recent DeepSeek model family \cite{deepseekai2025}.

\paragraph{Pipeline components} Most (13/20) teams broke the task down into multiple steps - typically, a binary classification of the context sentence as idiomatic or literal, followed by the image ranking. Many teams introduced variations in processing for literal and idiomatic instances, with alterations to LLM prompts, input data augmentation (particularly for text passed to CLIP) and in some cases models finetuned for each sentence type.

\paragraph{Mixtures of Experts} Four teams used a mixture-of-experts approach to one or more elements of the task, employing a variety of model types and sizes or prompt variation to smooth out inconsistencies in the model outputs. 

\paragraph{Challenges of LLMs} Several teams reported that they took steps to try to ensure that the outputs of generative LLMs were useful, including by prompting to constrain generation and by post-processing to parse the text. 
Three teams working with LLMs observed a bias in these models towards treating potentially idiomatic expressions as idiomatic regardless of the context in which they occur. Other work has described similar challenges when using LLMs for idiomaticity detection \cite{phelps-etal-2024-sign,mi_rolling_2024,he2024investigating}. See also \citet{khoshtab_comparative_2025} for discussion of prompting methods for LLM figurative language processing.

\paragraph{Data Augmentation} Seven teams describe some kind of data augmentation step, including paraphrasing and backtranslation to generate variations for finetuning, the addition of information about the target compound (by distillation from LLMs) and image variations. Two teams (FJWU\_Squad and dutir914) generated additional training instances automatically using generative model pipelines. We hope that these datasets will be made public by the authors in support of future research efforts.

\subsection{Most Effective Approaches}

\subsubsection{Subtask A}

The system submitted by AlexUNLP introduced a transformation step between the sentence classification and image ranking stages, in which compounds detected as being idiomatic are replaced with compositional synonyms (\textit{dirty money} becomes \textit{illegal money}). This step is intended to bypass the VLM's tendency to favour literal interpretations of compounds. The USCS NLP T6 team also comment on this phenomenon, and share our hypothesis that this likely stems from the use of image-caption datasets for model training; it seems likely that humans employ idiomatic language less frequently when captioning images\footnote{Very frequent idiomatic expressions and things which are often photographed may be exceptions - we recommend asking an image generation model to picture a \textit{hen party}.}.
AlexUNLP also evaluated a large number of different language and vision model combinations, with an ensemble of several models yielding the best results. For Portuguese, multimodal models outperformed translating into English
.

The second-placed system from dutir914 also used an ensemble of outputs from several QWEN2.5 models, with chain-of-thought prompting influencing the model generations. The team also generated additional training data, which was used to fine-tune CLIP. This data was produced by using DeepSeek to generate explanations and representative sentences for idiomatic expressions, which were passed to Flux.1-dev\footnote{\url{https://flux1ai.com/dev}} to generate images.

PALI-NLP obtained the highest overall performance for Subtask A. In particular, their methodology was the most successful on the extended evaluation sets for both English and Portuguese. Among other detailed adjustments to the prompts used to interact with LLMs, they introduce an interesting adjustment to counter the models' bias towards idiomatic interpretations. By first asking the LLM to produce examples of the expression used literally before providing the context sentence, they improve classification accuracy from 91.4 to 98.6\%. 
Examplars are also incorporated into the instruction prompts, with challenging ones purposefully selected. Multiple prompt variations are used, with the final output derived from aggregating the corresponding outputs \cite{wang2023scr}.

\subsubsection{Subtask B}

Subtask B (image sequence completion; \S\ref{sec:subtaskB_desc}) attracted few submissions, and no teams participated only in this subtask.
Due to the limited size of the development and test datasets, measured system performance was somewhat volatile. The highest-ranked system (daalft) reported a substantial drop in accuracy on the extended test set. PALI-NLP's approach again exhibited greater stability here. Their methodology involved prompting an LLM to generate a continuation of the narrative represented in the two initial images, then selecting from the candidate images based on their appropriateness for this generated continuation. As with subtask A, they achieved improvements of around 10\% by aggregating across the output of several prompt variations \cite{wang2023scr}.

\subsubsection{Text-Only Methods}

While most participating teams focused on the text \& image configuration, a few also submitted versions of their systems which used only the provided image captions (\S\ref{sec:dataset}) as input. These teams generally reported that the image data was beneficial to their overall performance. Several teams reported that (automatically) translating the captions into Portuguese improved the classification and ranking performance of their multimodal language models.
Some teams, especially those working with smaller models \citep[e.g. SBERT; ][]{SBERT}, found that they needed to shorten the supplied captions through truncation or summarisation.

Two teams participated only in the text-only tracks. The Transformer25 team employed smaller, fine-tuned SBERT models for the image ranking task, but augmented the captions with information about the target item generated by a large pretrained GPT-4 model.
The most successful text-only approach (CTYUN-AI) also employed data augmentation using synonym replacement and backtranslation, and this team also finetuned QWEN \cite{QWEN} models to the AdMIRe task. Interestingly, they report that their experiments with knowledge distillation from GPT-4 did not provide performance uplift, and that the largest QWEN2.5-72B model was no more effective than its 32B-parameter version.

\section{Human Evaluation}
\label{sec:human_eval}


In order to provide a comparison point with the model-driven systems, we presented subtask A (using the English extended evaluation dataset) to human annotators. 
Twelve (self-described) fluent speakers of English were recruited from among the staff and postgraduate research students of a UK University, and were compensated with a voucher to the value of GBP15 for their time. The annotations were collected using a survey deployed on the Qualtrics online platform\footnote{\url{https://www.qualtrics.com/}}. 
For each item, annotators were asked whether they are familiar with its idiomatic meaning. If they answered `Yes', they were then asked to use drag-and-drop to rank the candidate images according to how well they represent the meaning of the expression as it is used in the context sentence.

Each annotator saw between 50 and 75 of the 100 items in the extended evaluation dataset, selected at random, and each item was annotated 7-9 times.
Table \ref{tab:humanresults} shows the mean top 1 accuracy and DCG score across the human annotators on the dataset. We also include the metrics for the best-performing individual annotator and the results obtained by treating all of the annotators for each item as a pool of experts, ranking the images according to the mean rank assigned by the annotators.


\begin{table}[!htp]\centering
\caption{\centering Human annotator performance on the English extended evaluation set}\label{tab:humanresults}
\begin{tabular}{lrrr}\toprule
&Top 1 Acc &DCG Score \\\midrule
Annotator average &0.71 &3.22 \\
Best individual &0.86 &3.41 \\
Pool of experts &0.83 &3.39 \\
\bottomrule
\end{tabular}
\end{table}

These results would place the average annotator in 5th position against the systems submitted to the shared task on a leaderboard based on the extended evaluation set scores. The best individual annotator would outperform the model-driven systems, and the pool of experts approach would tie with the most performant system.

\section{Discussion}

\paragraph{Subtask B} We received few entries for subtask B. This may have been influenced by the smaller quantity of training data available, or perhaps the more complex semantics of the compounds meant participants opted to focus on subtask A in the first instance. The best-performing system for subtask B obtained impressive results, which suggests that the task may not be more difficult in practice.

\paragraph{Extended evaluation} The extended evaluation datasets appear to have yielded more robust results than the smaller test sets; models which were tuned on the training data did not always hold up very well on the larger test set, suggesting that some overfitting may have occurred. In this instance, there may be an advantage to have overlap between the compounds included in training and test sets, especially when they are used in different senses.

\paragraph{Languages} We were pleased to see the Portuguese datasets receiving attention from most of the participating teams. There was a smaller difference in performance between the English and Portuguese portions of the dataset than we were anticipating \cite{phelps-etal-2024-sign}, suggesting that multilingual LLMs' understanding of figurative elements of languages other than English may be improving.

\paragraph{LLMs} Participating teams, including the best-performing system overall, were able to get good results from large-scale LMs. However, this required substantial editing and refinement of both input prompts and generated output, and often the use of several parallel queries to smooth out model variance. These, presumably, came with corresponding costs in terms of human effort, money and computation (with its associated environmental impact). Most of the LLMs used by participants were also closed-source, making them difficult to examine in depth.

\paragraph{Human performance} The results of our human evaluation (\S\ref{sec:human_eval}) suggest that the task is by no means trivial for humans who are familiar with the expression in question, but that the expected order of the images we generated is reasonably well-aligned with human understanding.

\paragraph{Mixtures of `experts'} Mixture-of-experts approaches proved useful to increase  overall accuracy across various model scales. This suggests that individual language models may exhibit good `understanding' of particular idiomatic expressions, or be able to handle them in specific senses and contexts, but that no one model has a complete grasp on the phenomenon of idiomaticity. Generative LLMs also appeared to be inconsistent in their outputs, varying in response to how inputs are formulated and/or with their stochastic outputs. 
To some extent, these observations also hold for our human annotators, with no individual's responses perfectly matching the expected answers.

\section{Conclusion}

The AdMIRe shared task has encouraged participants to push the boundaries of idiomatic language representation by leveraging multimodal approaches incorporating both static and temporal visual cues. By expanding on previous idiomaticity-related tasks, AdMIRe introduces a dataset of nominal compounds that are both plausible and imageable in literal and figurative contexts, covering both English and Brazilian Portuguese. This challenge has provided valuable insights into the capabilities and limitations of contemporary vision-language models in processing idiomatic expressions.

While top-performing models achieved human-level performance, this success required significant computational resources and fine-tuning, which points to the limitations of these models. There remains considerable room for improvement in both the quality of idiomatic understanding in language models and their efficiency in processing these expressions. Future iterations of the dataset and task could drive further advancements by incorporating more languages, diverse figurative expressions, and refined methodologies that minimize dataset artifacts \citep{boisson_constructionartifacts}.

Ultimately, the AdMIRe challenge contributes to the ongoing effort to bridge the gap between human and machine language comprehension. This fosters progress in NLP applications such as machine translation, sentiment analysis, and the broader goal of understanding language.





\section*{Limitations}

\paragraph{Dataset size} The datasets for both subtasks are small in the context of contemporary machine learning data, which limited how much they could be used to train or fine-tune models. However, we believe that the manual curation of target items, context sentences, image prompts and generated images by native speakers means that the data quality is high. Individual idiomatic expressions are encountered infrequently, even in well-resourced languages such as English, yet are generally well-understood by human readers; we consider the AdMIRe dataset to represent a realistic yet achievable challenge for NLP systems.

\paragraph{Language variety} The AdMIRe dataset contains items in only two languages, both of which are relatively well-resourced. Expanding the dataset to a greater diversity of languages could enable better evaluation of idiomaticity representation.

\paragraph{Cultural background} The datasets were constructed by creators who are primarily middle-class and work in academic settings, and who are speakers of Brazilian Portuguese and/or British English. The same applies to most of the human annotators whose responses we used for evaluation. Our backgrounds and experiences will certainly have influenced the idiomatic expressions and context sentences we selected, the visual representations we favoured and so on.

\paragraph{AI tools used} Similarly, the datasets are likely to reflect biases and limitations present in the tools we used to construct them, especially the image generation and captioning models. While we made efforts to introduce diversity in the people depicted, we encountered some challenges in this regard. For instance, we were unable to identify prompts which would successfully generate images depicting a literal \textit{one-armed bandit}\footnote{Idiomatically, a slot machine.}. Expressions like \textit{blue blood} also presented difficulty, and we note that the depictions of female characters in particular in the dataset tend to be `conventionally attractive' and have somewhat exaggerated facial features.

\newpage

\section*{Acknowledgments}

The authors would like to acknowledge the contributions of our Brazilian collaborators who made this work possible. In particular, our thanks go to Rozane Rebechi, Juliana Carvalho, Eduardo Victor, César Rennó Costa, Marina Ribeiro and their colleagues and students.

We would also like to thank the members of the Multi-Word Expressions Group for their continued feedback and support. 
\vspace{1em}

This work was supported by the UKRI AI Centre for Doctoral Training in Speech and Language Technologies (SLT) and their Applications funded by UK Research and Innovation [grant number EP/S023062/1]. For the purpose of open access, the author has applied a Creative Commons Attribution (CC BY) licence to any Author Accepted Manuscript version arising.

This work also received support from the CA21167 COST action UniDive, funded by COST (European Cooperation in Science and Technology).

This work was supported by the EQUATE project.

\newpage

\bibliography{anthology,custom}


\appendix

\end{document}